\documentclass[letterpaper]{article} 
\usepackage{aaai25}  
\usepackage{times}  
\usepackage{helvet}  
\usepackage{courier}  
\usepackage[hyphens]{url}  
\usepackage{graphicx} 
\usepackage{natbib}  
\usepackage{caption} 
\usepackage{amssymb}

\usepackage{algorithm}
\usepackage{algorithmic}
\usepackage{amsmath} 
\usepackage{booktabs} 
\usepackage{booktabs} 
\usepackage{multirow} 
\usepackage{amssymb} 
\usepackage{graphicx} 
\usepackage[table]{xcolor}

\usepackage{newfloat}
\usepackage{listings}
\DeclareCaptionStyle{ruled}{labelfont=normalfont,labelsep=colon,strut=off} 

\floatstyle{ruled}
\newfloat{listing}{tb}{lst}{}
\floatname{listing}{Listing}
\title{Incomplete Modality Disentangled Representation \\ for Ophthalmic Disease Grading and Diagnosis}
\author{
    Chengzhi Liu\textsuperscript{\rm 1,2}\equalcontrib,
    Zile Huang\textsuperscript{\rm 1,2}\equalcontrib,
    Zhe Chen\textsuperscript{\rm 2},
    Feilong Tang\textsuperscript{\rm 2}\textsuperscript{\dag},
    Yu Tian\textsuperscript{\rm 4},
    Zhongxing Xu\textsuperscript{\rm 2}, \\
    Zihong Luo\textsuperscript{\rm 2},
    Yalin Zheng\textsuperscript{\rm 2},
    Yanda Meng\textsuperscript{\rm 1,2}\thanks{Corresponding author: Feilong Tang; Yanda Meng}
}
\affiliations{
    \textsuperscript{\rm 1} Department of Computer Science, University of Exeter, UK \\
    \textsuperscript{\rm 2}
    Department of Eye and Vision Sciences, University of Liverpool, UK \\
    \textsuperscript{\rm 4} Center For AI And Data Science For Integrated Diagnostics, University of Pennsylvania, USA\\
    Y.M.Meng@exeter.ac.uk
}

\usepackage{bibentry}

\begin{document}

\maketitle

\begin{abstract}
Ophthalmologists typically require multimodal data sources to improve diagnostic accuracy in clinical decisions. However, due to medical device shortages, low-quality data and data privacy concerns, missing data modalities are common in real-world scenarios. Existing deep learning methods tend to address it by learning an implicit latent subspace representation for different modality combinations. We identify two significant limitations of these methods: (1) implicit representation constraints that hinder the model's ability to capture modality-specific information and (2) modality heterogeneity, causing distribution gaps and redundancy in feature representations. To address these, we propose an Incomplete Modality Disentangled Representation (IMDR) strategy, which disentangles features into explicit independent modal-common and modal-specific features by guidance of mutual information, distilling informative knowledge and enabling it to reconstruct valuable missing semantics and produce robust multimodal representations. Furthermore, we introduce a joint proxy learning module that assists IMDR in eliminating intra-modality redundancy by exploiting the extracted proxies from each class. Experiments on four ophthalmology multimodal datasets demonstrate that the proposed IMDR outperforms the state-of-the-art methods significantly.
\end{abstract}

\section{Introduction}
Retinal Fundus Imaging and Optical Coherence Tomography are widely used 2D and 3D imaging techniques for detecting ophthalmic diseases. Recent methods have integrated multiple modalities to enhance diagnostic accuracy~\cite{watanabe2022combining, wang2023fundus, peng2024sam, zou2023reliable, tang2024discriminating,trinh2024sight,xiong2024sam2,li2024tp,wang2023knowledge,duan2024towards,qu2023openal,meng2024continuous,peng2023usage}. Despite the benefits of more comprehensive diagnostic insights compared to single-modality approaches, factors such as low-quality data, lack of equipment, and data privacy concerns in real-world practice settings can lead to missing modalities~\cite{warner2024multimodal}. Developing a reliable method for ophthalmic disease diagnosis that effectively addresses incomplete modalities remains challenging.\par

\begin{figure*}[ht]
    \centering
\includegraphics[width=\linewidth]{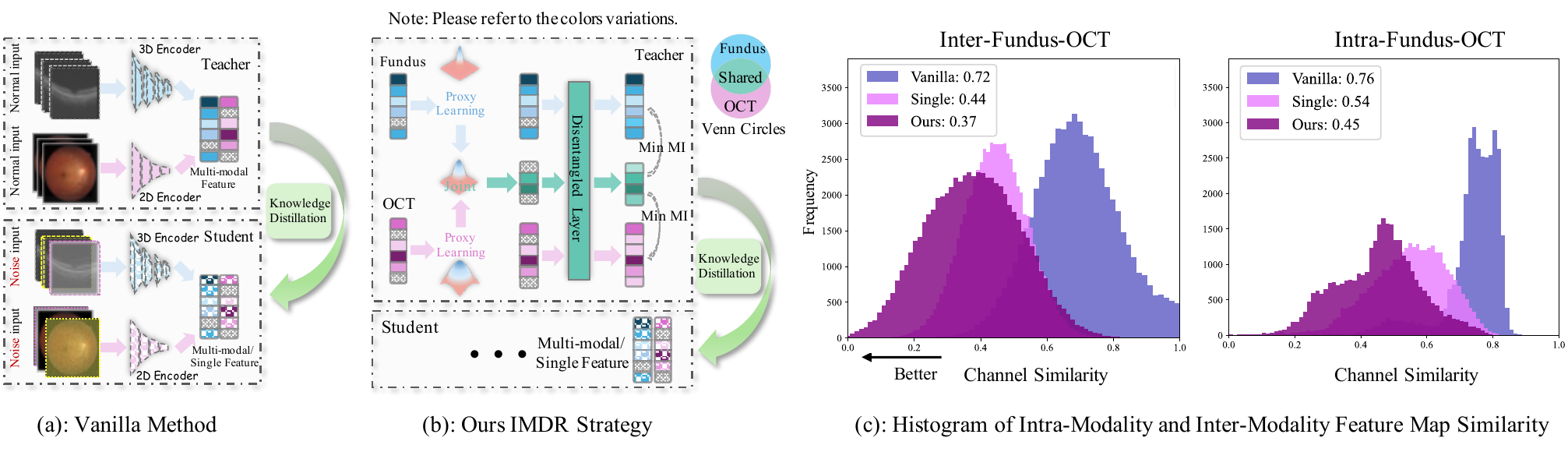}
    \caption{(a) Vanilla latent subspace methods. (b) Our proposed IMDR strategy effectively decouples multimodal data by employing explicit constraints to minimize mutual information in a Disentangle Extraction layer, guided by a joint distribution. (c) Illustration of intra-modality and inter-modality inter-channel distances between encoder feature maps. “Single” is the model that trains the encoders of each modality independently, providing the ideal feature diversity without inter-modality interference. “:A” denotes the histogram mean. Lower inter-channel similarity means higher diversity. More details in the Appendix B. }
    \label{fig1}
\end{figure*}

Existing methods for handling missing-modality problems can be classified into two categories: (1) Naive Modality Generation~\cite{ wang2023distribution, chen2024towards} reconstructs missing modalities from the available ones. However, controlling and generating medical image quality is complex due to a lack of clinical knowledge background, thus introducing noise and suffering from computational burden. 
(2) Latent Subspace Methods~\cite{liu2023m3ae, wang2023multi, liu2023sfusion, wang2023prototype, shi2024passion, li2024correlation, tang2024hunting,yang2024tackling} directly project inputs with different modality combinations into a deterministic embedding, ensuring consistency between the features and logits of the student and teacher models, as illustrated in Fig.~\ref{fig1} (a). Despite the success of these methods, they suffer from poor scalability and limited feature representations during knowledge distillation and do not intrinsically investigate the integration among various modalities.\par

Under the missing-modality setting, we identify a sub-optimal challenge in multimodal distillation: Although the teacher model generates more valuable feature representations than the student model, these representations are redundant and lack modality-specific information, resulting in ineffective distillation.
The reasons for this challenge are twofold:
\textbf{(i) Modality Heterogeneity:} The discrepancies in the inherent properties, statistical distribution, and structural characteristics of data across different modalities lead to a feature distribution gap and information redundancy during multimodal joint learning, which is confirmed by the empirical observation in~\cite{liang2022mind,udandarao2023sus, hu2024ophclip}. These discrepancies are further exacerbated under the missing-modality setting, leading to degraded model performance.
\textbf{(ii) Implicit Representation Constraint:}  
As shown in Fig.~\ref{fig1} (c), different input combinations from the same class are forced to learn the embeddings in the same direction of latent space, reducing the diversity of features. The lack of feature diversity results in sub-optimal distillation, affecting the model performance. 


\par

To this end, we propose an \textbf{I}ncomplete \textbf{M}odality \textbf{D}isentangled \textbf{R}epresentation (IMDR) strategy for ophthalmic disease diagnosis to disentangle features into explicit independent modal-shared and modal-specific features by the guidance of the modality joint distribution, distilling informative knowledge from the teacher model to the student network and enabling it to reconstruct valuable missing semantics and produce robust multimodal representations, as shown in Fig.~\ref{fig1} (b). 
Specifically, IMDR constructs probabilistic modality-specific representations obeying Gaussian distribution, allowing their combination into a joint distribution that estimates the latent space for each modality input. 
Modal-shared information is extracted by sampling from this distribution.
The Disentangle Extraction (DE) layer, using an attention mechanism, minimizes mutual information (MI) between modality mean representations, ensuring the features are effectively disentangled into independent modal-shared and modality-specific components, thus distilling informative knowledge to the student network.

To eliminate intra-modality redundancy and obtain a robust joint distribution, we introduce a Joint Proxy Learning (JPL) module that utilizes multiple sets of learnable proxies for each modality. Each set corresponds to a latent space distribution aligned with its ground truth label, effectively capturing discriminative information. The module reduces the overlap in feature representations between classes by maximizing the similarities between encoded single-modality features and positive proxies while minimizing those with negative proxies. Consequently, a joint distribution derived from proxies replaces the original joint distribution as the guidance, which becomes more refined and less redundant, enhancing the robustness of the model.

We evaluate the proposed method on intra and inter-modality incompleteness conditions across four ophthalmology multimodal datasets, where our approach achieves state-of-the-art performances. The contributions of our work are summarized as:
\begin{itemize}
\item We propose an Incomplete Modality Disentangled Representation (IMDR) strategy that disentangles features into explicit independent modal-shared and modal-specific features and distills informative knowledge from the teacher model to the student network.
\item We propose an auxiliary Joint Proxy Learning (JPL) module to eliminate intra-modality redundancy to obtain a robust joint distribution.
\item Experiments on four ophthalmology multimodal datasets demonstrate the effectiveness of the IMDR strategy.
\end{itemize}

\section{Related Work}
\subsubsection{Incomplete Multimodal Learning.} 
Recent methods in incomplete multimodal learning emphasize data reconstruction and latent subspace methods. Traditional data reconstruction methods utilize Generative Adversarial Networks~\cite{jue2019integrating,liu2021face} to enable models to simulate complete datasets but incur high computational costs. To address this, advanced techniques~\cite{li2023missing, miao2023multimodal, xiong2023client, poudel2024car, hu2025ophnet, xu2024polyp} reduce computational overhead. Additionally, statistical models have been introduced to align distributions of reconstructed and available data~\cite{zou2023reliable,wang2023distribution}. Latent subspace methods map different data modalities into a shared latent space, enabling the learning of intra-modality relationships~\cite{ebrahimi2023lanistr,sun2024redcore,wang2023distribution}. This work employs explicit constraints to remove redundant information and capture discriminative features, hence relaxing the implicit representation constraint.

\begin{figure*}[!ht]
\centering
\normalsize
\includegraphics[width=0.95\linewidth]{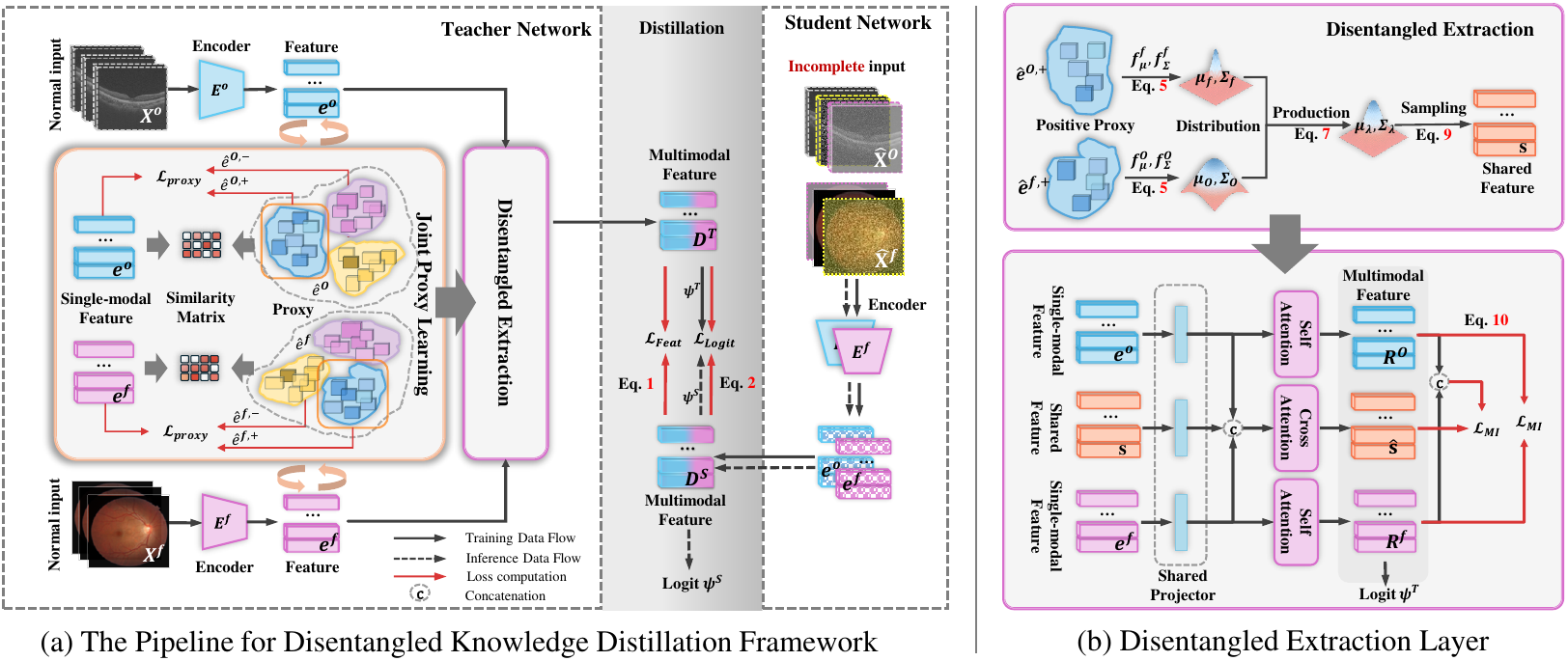}
    \caption{Overview of our proposed framework. \textbf{(a)}: 
    We train a teacher model using complete modality data, followed by co-training with a student model on incomplete inputs for knowledge distillation. The distillation is supervised by feature loss $\mathcal{L}_{\text{Feat}}$ and logit loss $\mathcal{L}_{\text{Logit}}$.
    During the training of the teacher model, the encoder outputs the single-modality feature $e^f$ and $e^O$.  We build a set of proxies for a modality, with each set representing a class. Positive proxies are selected by a similarity matrix between $\hat{e}$ and $e$. All proxies are optimized through the proxy loss $\mathcal{L}_{\text{Prox}}$. Consequently, $\hat{e}^{f,+}$ and $\hat{e}^{O,+}$, together with features $e^f$ and $e^O$ are then passed to the IMDR.
    \textbf{(b)}: Details for IMDR strategey. 
    We estimate the distributions of $\hat{e}^{f,+}$ and $\hat{e}^{O,+}$, then combine them using Eq.~\ref{eq: poe} to obtain the joint distribution $\mathcal{P}(\hat{e} \vert  x^{f}, x^{O})$. The modality-shared feature $s$ is sampled from this distribution. This feature $s$ guides the decoupling via an attention layer, supervised by the loss $\mathcal{L}_{\text{MI}}$ to minimize the mutual information between extracted shared features $\hat{s}$ and specific features $(\mathcal{R}^f, \mathcal{R}^O)$, as well as between $\mathcal{R}^f$ and $\mathcal{R}^O$.
    }
    \label{fig:enter-label}
\end{figure*}
\noindent\textbf{Decoupled Multimodal Representation.}
Decoupled representation techniques can be categorized into Modality-Specific Learning and Training-Inference Decoupling, which are vital for optimizing performance across diverse data types and operational phases.
Modality-Specific Learning ensures that the features of each data type are processed in a way that maintains their unique characteristics while being effectively integrated into a unified model~\cite{tripathi2024multimodal,li2023decoupled,cong2024decoupled, zhao2024sfc}. 
Training-Inference Decoupling involves using different models during the training phase compared to the inference phase to better adapt to varying task demands~\cite{zhang2024dept, tang2023duat, yang2023action, peng2024parameter}. 
Although these single-modality-based methods achieved remarkable improvements in feature extraction, they do not explicitly remove redundant information from inter-modality or intra-modality features.
This work utilizes joint distribution across modalities to further guide disentangling multimodal data while preserving diversity in modality shared and specific representations.

\section{Methodology}

\subsection{Problem Formulation}

We denote $X = \{x_i, y_i\}_{i=1}^N$ as a multimodal dataset that has $N$ samples. Each $x_i$ consists of $M$ inputs from different modalities as $x_i = \{ x_i^{m} \}_{m=1}^{M}$ and $y_i \in \{1, 2, \ldots, C\}$, where $M$ is the number of modalities; $C$ is the number of categories. The modality encoder $E$, learns from the input image $x$ to produce a single-modal representation $e$. The incomplete modality is denoted as $\hat{X}$, and we define two cases of incomplete modalities to simulate the natural and holistic challenges in real-world scenarios: \textbf{(i) intra-modality incompleteness}, referring to impaired or noisy data within a specific modality. \textbf{(ii) inter-modality incompleteness}, where some modalities are entirely missing, such as the frame-level features in the OCT layers. We use the scenario of intra-modality incompleteness as an example for clarity of illustration, as shown in Fig.~\ref{fig:enter-label}.

\subsection{Overall Framework}

\noindent\textbf{Training and Inference.} We first train the teacher model on complete-modality data $X$, and co-train with a student model with incomplete inputs $\hat{X}$ for knowledge distillation. Our goal is to model both intra- and inter-modality relations to create more informative multimodal features $D^T$ for distillation, enabling student model to reconstruct information in features $D^S$ during inference and obtain more accurate logits $\psi_S$ for ophthalmic disease grading and diagnosis across any degree of incomplete modalities.

\subsubsection{Teacher Model Training.} To eliminate intra-modality redundancy, we establish a set of proxies $\hat{e}$ to capture discriminative information relevant to the label $y$ for each modality. 
We optimize the parametric proxy $\hat{e}$ to align the distribution of the data $x$. 
By calculating the similarity between sets of $e$ and $\hat{e}$, we can identify a positive set of proxies $\hat{e}^{+}$. For better stability, we predict the mean $\mu$ and covariance matrix $\Sigma$ of proxies distribution by the distribution predictor $f_\mu$ and $f_\Sigma$, rather than modeling $\mathcal{P}(\hat{e} \vert x)$. The joint distribution $\mathcal{P}(z \vert  x^{f}, x^{O})$ is then obtained by multiplying the distributions of the positive proxies, and we randomly sample the modality-shared feature $s$ from this joint distribution.
Utilizing the modality-shared feature $s$ as guidance, we disentangle multimodal data into independent modality-shared representations $\hat{s}$ and modality-specific representations $\mathcal{R}^f$ and $\mathcal{R}^O$ for fundus and OCT within a Disentangle Extraction layer.
Specifically, a parameter-shared projector first obtain the query, key and value of $s$, $e^f$, and $e^O$, respectively. Consequently, modality-shared feature $\hat{s}$ is extracted by a plain cross-attention, which is detailed in Appendix D. $\mathcal{R}^f$ and $\mathcal{R}^O$ are extracted by self-attention module. A constraint is set to minimize the mutual information between the mean representations of $\hat{s}$, $\mathcal{R}^f$ and $\mathcal{R}^O$. Finally, $\hat{s}$, $\mathcal{R}^f$ and $\mathcal{R}^O$ are concatenated as multimodal feature $\mathcal{D}^T$ to obtain the predicted logits $\psi_T$ through a task-specific classifier. 

\subsubsection{Co-training Distillation.}
We employ feature-based and logit-based distillation to ensure consistency between the teacher model and the student model, enabling the student model to handle various scenarios of incomplete modality. 

In feature-based distillation, for each class $c$, a softmax loss is applied to the multimodal features to generate soft targets, denoted as $\Gamma_c^T = softmax(D_c^T / \tau)$, where $\tau$ is the temperature parameter that controls the softening of the distribution. The class deterministic feature $\Gamma_c^S$ of the student model is obtained similarly. The Kullback-Leibler (KL) divergence is then utilized to align the class deterministic feature of the student with those of the teacher, defined as:
\begin{equation}
\mathcal{L}_{\text{Feat}} = \sum_{c=1}^C \Gamma_c^T \log \left( \frac{\Gamma_c^T}{\Gamma_c^S} \right).
\end{equation}
In logit-based distillation, given the logits $\psi^T = \{\psi^T_c\}_{c=1}^C$ from the teacher model and $\psi^S$ from the student model, we enforce consistency between their cosine similarity matrices to effectively transfer inter-class relationship information. The similarity matrix for the teacher model is computed as $\mathcal{Q}_c^T = softmax(\text{sim}(\psi_c^T, \psi^T)/\tau)$, and similarly for the student model, $\mathcal{Q}_c^S$. The alignment between these similarity matrices is achieved through the KL divergence loss:
\begin{equation}
\mathcal{L}_{Logit} = \sum_{c=1}^C \mathcal{Q}_c^T \log \left( \frac{\mathcal{Q}_c^T}{\mathcal{Q}_c^S} \right).
\end{equation}
The comprehensive loss function integrates the classification loss and the distillation losses.
\begin{equation}\small
\mathcal{L}_{\text{Distill}} = \mathcal{L}_{\text{CE}} + \mathcal{L}_{\text{Feat}} + \mathcal{L}_{\text{Logit}},
\end{equation}
where $\mathcal{L}_{\text{CE}}$ is a cross-entropy loss for ophthalmic disease diagnosis task, allowing the same input to be associated with multiple classes.

\subsection{Distribution-guided Disentangling Strategy}
We introduce an IMDR strategy to effectively capture both modality-shared and modality-specific representations across input combinations from different modalities while eliminating inter-modality redundancy.

We consider building probabilistic embeddings, i.e. $e \sim \mathcal{P}(e\vert x)$, to achieve a more adaptable representation space for modality-specific representations. Specifically, we define probabilistic embeddings $e$ following a spherical Gaussian distribution as common. Since computing the posterior distribution is intractable, we variationally approximate the posterior distribution $\mathcal{P}(e\vert x)$ as:
\begin{equation}\small
\mathcal{P}(e\vert x) \approx  q_{\theta}(e\vert x)= \mathcal{N}(e; \mu, \Sigma),
\label{variational approximation}
\end{equation}
where $q_{\theta}(e\vert x)$ is a variation approximation, and the mean $\mu$ and covariance matrix $\Sigma$ of the Gaussian distribution are predicted by distribution predictors $f_\mu$ and $f_\Sigma$. Unlike existing methods~\cite{guo2017calibration, wei2022mitigating} that estimate $\mu$ and $\Sigma$ for the logits after pooling, we directly estimate $\mu$ and $\Sigma$ for the feature $e$.
For each modality, the parameters $\mu$ and $\Sigma$ are defined as:
\begin{equation}\small
\mu = f_\mu(\theta_\mu, e), \quad \log(\Sigma) = f_\Sigma(\theta_\Sigma, e),
\end{equation}
where $\theta_\mu$ and $\theta_\Sigma$ are the parameters for the distribution predictors $f_\mu(\cdot)$ and $f_\Sigma(\cdot)$, respectively. To enhance stability, we directly predict $\log(\Sigma)$ instead of $\Sigma$. Both $f_\mu(\cdot)$ and $f_\Sigma(\cdot)$ are implemented using a simple Multilayer Perceptron with batch normalization, introducing minimal additional parameters.

To extract modality-shared information, we leverage the joint distribution $\mathcal{P}(e\vert x^{f}, x^{O})$ across multiple modalities. This joint posterior is derived using the product-of-experts~\cite{hinton2002training}, where individual distributions are combined multiplicatively to form a cohesive model. The joint distribution $\mathcal{P}(e \vert  x^{f}, x^{O})$ can be formulated to: 
\begin{equation}\small
               \mathcal{P}(e\vert x^{f}, x^{O}) \propto \mathcal{P}(e)\mathcal{P}(e\vert x^{f})\mathcal{P}(e\vert x^{O}) ,
\end{equation}
where $\mathcal{P}(e\vert x^{f})$ and $\mathcal{P}(e\vert x^{O})$ could be approximated by Gaussian distributions $\mathcal{N}(e; \mu^{f}, \Sigma^{f})$ and $\mathcal{N}(e; \mu^{O}, \Sigma^{O})$ , as detailed in Eq.~\ref{variational approximation}. 
Since we define that the probabilistic representation $e$ follows a spherical Gaussian distribution, as proved in~\cite{zhang2024prototypical}, we can assume the prior distribution $\mathcal{P}(e)$ is also a spherical Gaussian $\mathcal{N}(e; \mu_\Delta, \Sigma_\Delta)$. Hence it can be shown that the product of Gaussian distributions is also a Gaussian distribution:
\begin{equation}\small
\begin{aligned}
\Sigma_\lambda &= (\Sigma_\Delta^{-1} + \sum_{m \in \{f,O\}} \Sigma_m^{-1})^{-1},\\ \quad \mu_\lambda &= (\mu_\Delta \Sigma_\Delta^{-1} + \sum_{m \in \{f,O\}} \mu_m \Sigma_m^{-1}) \Sigma_\lambda^{-1},
\end{aligned}
\label{eq: poe}
\end{equation}
which yields the parameters $\mu_\lambda, \Sigma_\lambda$ for joint distribution formulated as a Gaussian distribution:
\begin{equation}\small
    \mathcal{P}(e\vert x^{f}, x^{O}) = \mathcal{N}(e; \mu_\lambda, {\Sigma_\lambda}),
\end{equation}
hence, the modality-shared feature $s$ becomes a stochastic embedding sampled from the Gaussian distribution $\mathcal{N}(e; \mu_\lambda, {\Sigma_\lambda})$. To enable differentiable sampling, we employ the reparameterization trick~\cite{kingma2013auto}. In general, we sample noise from $\mathcal{N}(0, \mathbf{I})$ and construct the embedding $s$ by reparameterizating, instead of directly sampling from $\mathcal{N}(e; \mu_\lambda, \Sigma_\lambda)$:
\begin{equation}\small
s = \mu_\lambda + \epsilon \Sigma_\lambda, \quad \epsilon \sim \mathcal{N}(0, \mathbf{I}).
\label{sample}
\end{equation}

We further disentangle multimodal data into ideally independent modal-shared representations $\hat{s}$ and modal-specific representations $\mathcal{R}^{f}$, $\mathcal{R}^{O}$ through an attention layer. Then, we minimize the mutual information (MI) between modal-shared and modal-specific representations to preserve modality-specific information via a defined loss function:
\begin{equation}\small
\mathcal{L}_{\text{MI}} = I(\hat{s}, \Tilde{\mathcal{R}}) + I(\mathcal{R}^{f}, \mathcal{R}^{O}),
\end{equation}
where $I(\cdot, \cdot)$ represents the MI that measures the dependence between two variables, and $\Tilde{\mathcal{R}}$ is the concatenation of $\mathcal{R}^{f}$ and $\mathcal{R}^{O}$ as complete modal-specific information. Since MI is generally intractable, we employ CLUB~\cite{cheng2020club} to implement $\mathcal{L}_{\text{MI}}$. Detailed computation procedures are provided in the Appendix C.

\subsection{Joint Proxy Learning Module}
We introduce a JPL module to mitigate redundancy in modality-shared features from interfering with the disentanglement process.
For each modality, the module directly approximates the distribution $\mathcal{P}(e\vert x)$ with the distribution of a parametric proxy $\hat{e}$, with distribution $\mathcal{P}(\hat{e})$ represented by a set of $N_p$ proxies for each class, denoted as $\mathcal{P} = \{\mathcal{N}(\hat{e}; \mu_c, \Sigma_c)\}_{c=1}^{C}$. Each proxy is designed to capture discriminative information for the ground truth label $y$, approximating the conditional probability distribution $\mathcal{P}(\hat{e}\vert y) = \mathcal{N}(\hat{e}; \mu_y, \Sigma_y)$. The feature representations $e$ are then expected to align with the proxies $\hat{e}$ with the same label $y$. Consequently, the objective for the variation approximation in Eq.~\ref{variational approximation} is adjusted as:
\begin{equation}\small
\mathcal{P}(e \vert x) = \mathcal{P}(e \vert x, y) \approx \mathcal{P}(\hat{e} \vert y).
\label{approx proxy}
\end{equation}

To achieve this, we maximize the similarity between $\mathcal{P}(\hat{e})$ and distributions of features $e$. As a result, we just need to optimize the parametric proxies $\hat{e}$. The objective of approximating $\mathcal{P}(e\vert x,y)$ with proxies $\hat{\mathcal{P}}(\hat{e}\vert y)$ in Eq.~\ref{approx proxy} can be achieved by a proxy loss. For each modality $m$, the loss pulls the feature $e^m$ closer to positive proxies $\hat{e}^{m,+}$ while pushing them away from negative ones $\hat{e}^{m,-}$, formulated as:
\begin{equation}\scriptsize
    \mathcal{L}_{\text{Prox}} = \frac{1}{M}\sum_{m=1}^{M} \left ( -\text{Sim}(\hat{e}^{m,+}, e^{m}) + \frac{1}{C-1} \sum_{n=1}^{C-1} \text{Sim}(\hat{e}^{m,-}_n, e^{m}_n) \right ).
\end{equation}
During training, positive proxies $\hat{e}^{+}$ are selected based on their matching labels with the input data. During inference, this is determined by calculating a similarity matrix and selecting the set with the highest mean similarity.

Now that we can obtain $\hat{\mu}$ and $\hat{\Sigma}$ using positive proxies as inputs to the distribution predictor. We reformulated the distribution prediction as:
\begin{equation}\small
\hat{\mu} = f_\mu(\theta_\mu, \hat{e}^{+}), \quad \log(\hat{\Sigma}) = f_\Sigma(\theta_\Sigma, \hat{e}^{+}).
\end{equation}
Hence, with $\mathcal{P}(\hat{e})$ being a spherical Gaussian $\mathcal{N}(\hat{e}; \mu_\Delta, \Sigma_\Delta)$ the $\mu_\lambda$ and $\Sigma_\lambda$ for joint distribution defined in Eq.~\ref{eq: poe} can be reformulated to: 
\begin{equation}\small
\begin{aligned}
\hat{\Sigma}_\lambda &= (\Sigma_\Delta^{-1} + \sum_{m \in \{f,O\}} \hat{\Sigma}_m^{-1})^{-1},\\ 
\quad \hat{\mu}_\lambda &= (\mu_\Delta \Sigma_\Delta^{-1} + \sum_{m \in \{f,O\}} \hat{\mu}_m \hat{\Sigma}_m^{-1}) \hat{\Sigma}_\lambda^{-1},
\end{aligned}
\end{equation}
which leads to the joint distribution $\mathcal{P}(\hat{e}\vert x^{f},x^{O})$:
\begin{equation}\small
\mathcal{P}(\hat{e} \vert x^f, x^O) = \mathcal{N}(\hat{e}; \hat{\mu}_\lambda, \hat{\Sigma}_\lambda).
\end{equation}

The overall learning objective for the teacher model is:
\begin{equation}\small
\mathcal{L}_{\text{Teacher}} = \mathcal{L}_{\text{CE}} + \omega_1 \mathcal{L}_{\text{MI}} + \omega_2 \mathcal{L}_{\text{Prox}}.
\end{equation}
where $\omega_1$ and $\omega_2$ are weights that control the contribution of
the mutual information and proxy losses, respectively.

\section{Experiments}
\subsection{Datasets and Evaluation Metrics}
We evaluate the proposed framework using four publicly available multimodal datasets:  \textbf{GAMMA} dataset~\cite{WU2023102938} and three subsets from Harvard-30k~\cite{luo2024eye}, including \textbf{Harvard-30k AMD}, \textbf{Harvard-30k DR}, and \textbf{Harvard-30k Glaucoma}, covering Age-related Macular Degeneration (AMD), Diabetic Retinopathy (DR), and Glaucoma. GAMMA contains cases with three-tier grading, with OCT and fundus images sized at 256 × 512 × 992 and 1956 × 1934, respectively. Harvard-30k subsets are annotated with four-tier AMD grading and two-tier for glaucoma and DR, with image dimensions of 448 × 448 for fundus and 200 × 256 × 256 for OCT (where 200 denotes the number of OCT slices). Detailed descriptions are in Appendix E.

To ensure reliable results, each dataset underwent five-fold cross-validation, and the model is assessed using four key metrics: Accuracy (ACC), F1 score (F1), Area Under the Curve (AUC), and Specificity (Spec), effectively predicting the severity and type of various ophthalmic diseases.

\subsection{Comparison with State-of-the-art Methods}
\subsubsection{Effectiveness in Complete-modality Fusion.} 
We compare IMDR with six representative complete-modality fusion methods using CNN and Transformer architectures, as shown in Table~\ref{tab:Quantitative-Results}. ResNet50 2D and 3D serve as CNN backbones, while Swin-Transformer and UNETR are employed as Transformer backbones. \textbf{(1) B-IF}, a baseline utilizing an intermediate multimodal fusion method; \textbf{(2) HFS-IL}, which employs a hybrid fusion strategy that combines intermediate and late fusion;
\textbf{(3) B-EF}~\cite{hua2020convolutional}, an early fusion strategy; 
\textbf{(4) CR-AF}~\cite{zheng2023casf}, incorporating a multimodal cross-attention fusion approach; \textbf{(5) $\text{M}^2$LC}~\cite{woo2018cbam}, integrating both channel attention and spatial attention for fusion; \textbf{(6) Eye-Most}~\cite{ZOU2024103214}, an evidence fusion model based on the inverse gamma prior distribution. The Complete-Modality Fusion section of Table~\ref{tab:Quantitative-Results} shows that the proposed IMDR model consistently outperforms other methods across all datasets and backbones evaluated. For example, on the Harvard-30k AMD dataset, our IMDR model outperforms the state-of-the-art Eye-Most model by over +3.58\% in accuracy and exceeds the baseline model (B-IF) by +6.33\% in specificity when compared to the Transformer baseline architecture.

\begin{table*}[ht]
    \centering
    \resizebox{0.9\textwidth}{!}{%
    \begin{tabular}{c|cc|ccc|ccc|ccc|ccc}
        \toprule
        \multirow{2}{*}{\textbf{Method}} 
        & \multicolumn{2}{c|}{\textbf{Modality}} 
        & \multicolumn{3}{c|}{\textbf{GAMMA}} 
        & \multicolumn{3}{c|}{\textbf{Harvard-30k AMD}} 
        & \multicolumn{3}{c|}{\textbf{Harvard-30k DR}} 
        & \multicolumn{3}{c}{\textbf{Harvard-30k Glaucoma}} \\
        \cmidrule(r){2-15}
        & \textbf{OCT} & \textbf{Fundus} 
        & \textbf{ACC} & \textbf{AUC} & \textbf{F1} 
        & \textbf{ACC} & \textbf{AUC} & \textbf{F1} 
        & \textbf{ACC} & \textbf{AUC} & \textbf{F1} 
        & \textbf{ACC} & \textbf{AUC} & \textbf{F1} \\
        \midrule
        \multicolumn{14}{c}{\textbf{Inter-Modality Missing with ResNet-50 Backbone}} \\
        \midrule
        2D-Resnet50 Backbone & & $\textbf{\checkmark}$ & 0.7050 & 0.7896 & 0.6409 & 0.7312 & 0.7535 & 0.7223 & 0.7381 & 0.7915 & 0.7047 & 0.7312 & 0.7535 & 0.7223 \\
        3D-Resnet50 Backbone & $\textbf{\checkmark}$ & & 0.6860 & 0.7453 & 0.6210 & 0.6517 & 0.6998 & 0.6963 & 0.7073 & 0.6994 & 0.6197  & 0.6572 & 0.6989 & 0.7098 \\
         B-IF + distill &   &$\textbf{\checkmark}$ & 0.7153 & 0.8005 & 0.7233 & 0.7235 & 0.7198 & 0.7003 & 0.7362 & 0.6750 & 0.6968 & 0.7339 & 0.7661  & 0.7247\\
        B-IF + distill &  $\textbf{\checkmark}$ & & 0.6891 & 0.7859 & 0.6844 & 0.6957 & 0.7014 & 0.6745 & 0.6905 & 0.6525 & 0.6793 & 0.6964 & 0.6895 & 0.6718 \\
        $\text{M}^2$LC + distill &  & $\textbf{\checkmark}$ & 0.7607 & 0.7561 & 0.7548 & 0.7324 & 0.7267 & 0.7380 & 0.7304 & 0.6789 & 0.7459 & 0.7278 & 0.7023 & 0.7111 \\
        $\text{M}^2$LC + distill & $\textbf{\checkmark}$ & & 0.7205 & 0.7566 & 0.7275 & 0.6897 & 0.7223 & 0.6506 & 0.6720 & 0.6505 & 0.6433 &  0.6770 & 0.7122 & 0.6560 \\
        \rowcolor{gray!30} \textbf{IMDR (Ours)}   &  & $\textbf{\checkmark}$ & \textbf{0.7900} & \textbf{0.7655} & \textbf{0.7433} & \textbf{0.7517} & \textbf{0.8048} & \textbf{0.7659} & \textbf{0.7619} & \textbf{0.7907} & \textbf{0.7218} & \textbf{0.7544} & \textbf{0.7847} & \textbf{0.7512} \\
        \rowcolor{gray!30} \textbf{IMDR (Ours)} & $\textbf{\checkmark}$ &  & \textbf{0.7483} & \textbf{0.7912} & \textbf{0.8097} & \textbf{0.7062} & \textbf{0.7269} & \textbf{0.7190} & \textbf{0.7262} & \textbf{0.7469} & \textbf{0.7290} & \textbf{0.7116} & \textbf{0.7507} & \textbf{0.7037} \\
        \midrule
        \multicolumn{14}{c}{\textbf{Complete-Modality Fusion with ResNet-50 Backbone}} \\
        \midrule
        B-IF & $\textbf{\checkmark}$ & $\textbf{\checkmark}$ & 0.7099 & 0.8610 & 0.6691 & 0.7167 & 0.8122 & 0.6901 & 0.7355 & 0.7544 & 0.7396 & 0.7366 & 0.7937 & 0.7289 \\
        B-EF & $\textbf{\checkmark}$ & $\textbf{\checkmark}$ & 0.6837 & 0.8024 & 0.6433 & 0.6933 & 0.7912 & 0.6064 & 0.7498 & 0.7413 & 0.7717 & 0.7419 & 0.7752 & 0.7222 \\
        HFS-IL & $\textbf{\checkmark}$ & $\textbf{\checkmark}$ & 0.7431 & 0.8347 & 0.6800 & 0.7500 & 0.8174 & 0.7104 & 0.7557 & 0.7531 & 0.7647 & 0.7334 & 0.8096 & 0.7281 \\
        CR-AF & $\textbf{\checkmark}$ & $\textbf{\checkmark}$ & 0.6703 & 0.8123 & 0.6777 & 0.7300 & 0.8137 & 0.6773 & 0.7657 & 0.7835 & 0.7627 & 0.7392 & 0.7748 & 0.7309 \\
        $\text{M}^2$LC & $\textbf{\checkmark}$ & $\textbf{\checkmark}$ & 0.7200 & 0.8609 & 0.6804 & 0.7400 & 0.8456 & 0.7073 & 0.7736 & 0.8362 & 0.7601 & 0.7473 & 0.7745 & 0.7354 \\
        Eye-Most & $\textbf{\checkmark}$ & $\textbf{\checkmark}$ & 0.8600 & 0.8493 & 0.8135 & 0.7555 & 0.8230 & 0.7100 & 0.7500 & 0.8034 & 0.7342 & 0.7433 & 0.7540 & 0.7201 \\
        \rowcolor{gray!30} \textbf{IMDR (Ours)} & $\textbf{\checkmark}$ & $\textbf{\checkmark}$ & \textbf{0.8700} & \textbf{0.8320} & \textbf{0.8366} & \textbf{0.7645} & \textbf{0.8310} & \textbf{0.7357} & \textbf{0.7803} & \textbf{0.8534} & \textbf{0.7823}  & \textbf{0.7650} & \textbf{0.7711}  & \textbf{0.7437}\\
        \midrule
        \multicolumn{14}{c}{\textbf{Complete-Modality Fusion with ResNet-101 Backbone}} \\
        \midrule
        B-IF & $\textbf{\checkmark}$ & $\textbf{\checkmark}$ & 0.7100 & 0.8856 & 0.7076 & 0.7317 & 0.8382 & 0.7125 & 0.7636 & 0.7795 & 0.7561 & 0.7339 &  0.7332 &  0.7211\\
        HFS-IL & $\textbf{\checkmark}$ & $\textbf{\checkmark}$ & 0.7326 & 0.8106 & 0.7005 & 0.7217 & 0.8181 & 0.7196 & 0.7083 & 0.6119 & 0.6884 & 0.7415 & 0.7878 & 0.7681 \\
        CR-AF & $\textbf{\checkmark}$ & $\textbf{\checkmark}$ & 0.7425 & 0.8077 & 0.7562 & 0.7436 & 0.8274 & 0.7348 & 0.7489 & 0.7592 & 0.7581 & 0.7324 & 0.7593 & 0.7420 \\
        $\text{M}^2$LC & $\textbf{\checkmark}$ & $\textbf{\checkmark}$ & 0.7860 & 0.7800 & 0.6823 & 0.7493 & 0.8239 & 0.7120 & 0.7521 & 0.7968 & 0.7439 & 0.7498 & 0.7645 & 0.7423 \\
        Eye-Most & $\textbf{\checkmark}$ & $\textbf{\checkmark}$ & 0.8200 & 0.8321 & 0.7488 & 0.7592 & 0.8334 & 0.7002 & 0.7688 & 0.8259 & 0.7588 & 0.7210 & 0.7200 & 0.7342 \\
        \rowcolor{gray!30} \textbf{IMDR (Ours)} & $\textbf{\checkmark}$ & $\textbf{\checkmark}$ & \textbf{0.8775} & \textbf{0.8495} & \textbf{0.7958} & \textbf{0.7950} & \textbf{0.8509} & \textbf{0.7252} & \textbf{0.7857} & \textbf{0.8500} & \textbf{0.7704} & \textbf{0.7731} & \textbf{0.7898} & \textbf{0.7890} \\
        \bottomrule
    \end{tabular}%
    }
    \caption{Quantitative Results for Missing-Modality Tasks on GAMMA, Harvard-30k AMD, DR, and Glaucoma Datasets.}
    \label{tab:Quantitative-Results}
\end{table*}

\subsubsection{Robustness to Inter-modality Incompleteness.}
In the Inter-Modality Missing section of Table~\ref{tab:Quantitative-Results}, we evaluate our model by comparing the performance of the single backbone and other models. All models show a performance decline when a modality is missing, with the most pronounced drop occurring in the absence of fundus images. This highlights the critical role of complementary information from heterogeneous modalities. However, IMDR demonstrates greater robustness, retaining strong performance when the OCT modality is missing and surpassing other models when the crucial fundus modality is absent. This robustness is due to IMDR's capability to disentangle multimodal features, distill informative knowledge to the student network, and reconstruct missing semantics for robust representations.

\begin{figure}[t]
    \centering
\includegraphics[width=0.48\textwidth]{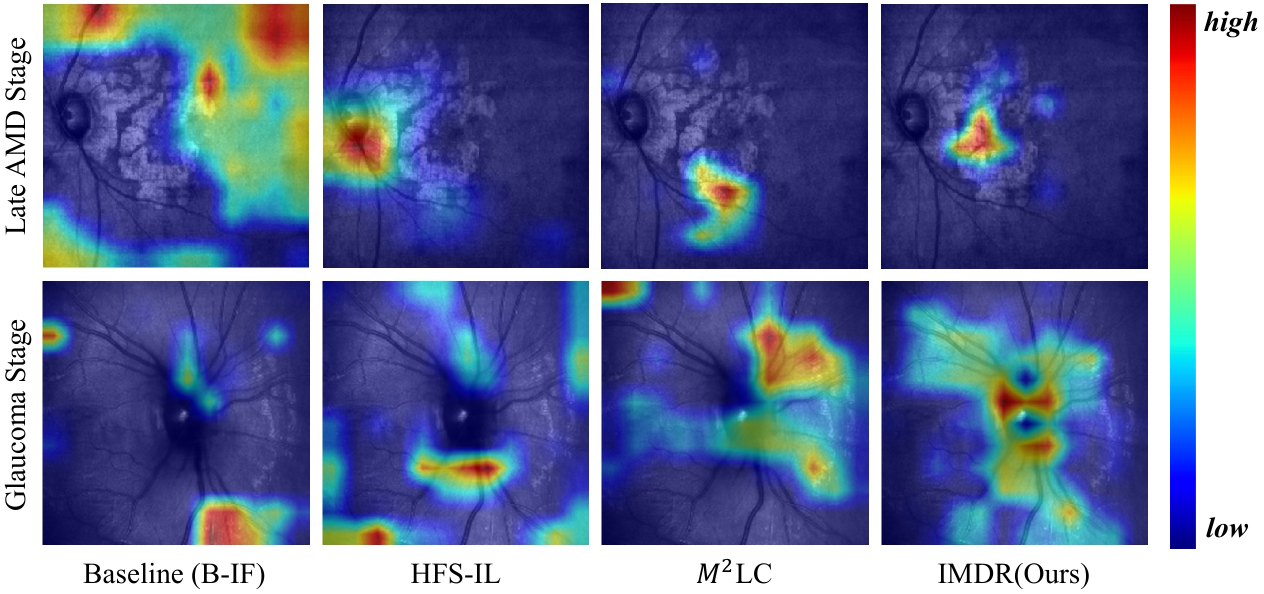}
    \caption{Comparative visualization of attention maps under the inter-modality incompleteness setting: The first row is AMD dataset and the second row is Glaucoma dataset.}
\label{cam}
\vspace{-0.5cm}
\end{figure}

As shown in Fig.~\ref{cam}, we compare GradCAM~\cite{selvaraju2017grad} heatmaps of IMDR with other models under missing OCT modality on AMD and Glaucoma datasets. Unlike baseline and other models, which miss critical regions, IMDR accurately captures essential features, such as optic nerve head information for Glaucoma and hemorrhage features for AMD, due to its effectiveness in capturing discriminative information relevant to the target.

\subsubsection{Robustness to Intra-modality Incompleteness.}  
\begin{figure*}[t]
    \centering
\includegraphics[width=0.96\textwidth]{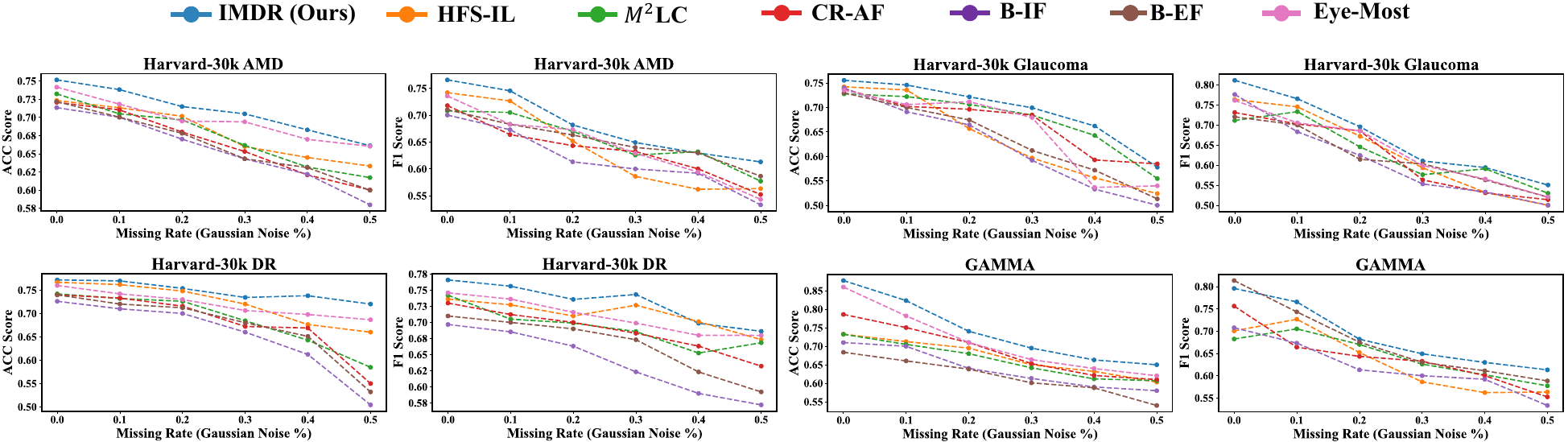}
    \caption{The comparison of performance across various missing rates under intra-modality incompleteness.}
    \label{fig:various_missing_rates}
\end{figure*}

We estimate the robustness of intra-modality incompleteness by comparing performance across four datasets under different data missing rates, as shown in Fig.~\ref{fig:various_missing_rates}.
To simulate intra-modality data loss, we introduce Gaussian noise at varying levels $(\alpha=[0.1, 0.5])$ to represent different degrees of information degradation. As noise increases, all models show significant performance declines, highlighting the impact of single-modality data loss on the stability of multimodal representations. However, our IMDR model exhibits superior robustness, particularly under high information loss, maintaining relatively stable performance compared to other models. This underscores the effectiveness of our proxy learning in capturing and modeling class relations, enabling student network to reconstruct missing information.

\begin{table}[t]
\centering
\normalsize  
\resizebox{0.42\textwidth}{!}{
\begin{tabular}{c|cccc|ccc}
\midrule
\textbf{Exp} & C-Feat & C-Dist & JPL & DE & \textbf{ACC} & \textbf{AUC} & \textbf{F1} \\ 
\midrule
I &  \textbf{\checkmark} & &  &  & 0.633 & 0.625 & 0.677  \\ 
II & & \textbf{\checkmark}  & &  & 0.654 & 0.643 & 0.623 \\ 
III & & \textbf{\checkmark} &  & \textbf{\checkmark} & 0.681 & 0.755 & 0.712  \\ 
IV & & \textbf{\checkmark} & \textbf{\checkmark} & & 0.725 & 0.672 & 0.680  \\ 
V & & \textbf{\checkmark} & \textbf{\checkmark} & \textbf{\checkmark} & \textbf{0.752} & \textbf{0.805} & \textbf{0.766} \\ 
\bottomrule
\end{tabular}
}
\caption{Ablation study with missing OCT modality.}
\label{intermiss_3d}
\vspace{-0.5cm}
\end{table}

\subsection{Ablation Studies}
\subsubsection{Effectiveness of each component.} 
To validate the effectiveness of each proposed component, we conduct five ablation experiments on the AMD dataset with the OCT modality missing, as shown in Table~\ref{intermiss_3d}. 
In Experiments I and II, C-Dist outperforms C-Feat, indicating that distribution-based methods effectively capture and integrate cross-modal representations. 
Experiment III shows that introducing our DE layer into the joint distribution, which preserves independent modality-shared and modality-shared features, increases accuracy by +2.7\%. Experiment IV, which incorporates our JPL module to eliminate redundancy, resulted in a +7.1\% performance boost. Experiment V combines the DE layer and the JPL module, further improving performance by a significant +9.8\% over Experiment II. This confirms that integrating our proposed components substantially enhances the robustness of the student model in the Ophthalmic Disease Diagnosis, demonstrating that the JPL and DE modules can work harmoniously together, with JPL capturing discriminative features and DE disentangling features through the joint distribution.

\begin{figure}[t]
\centering
\includegraphics[width=0.93\linewidth]{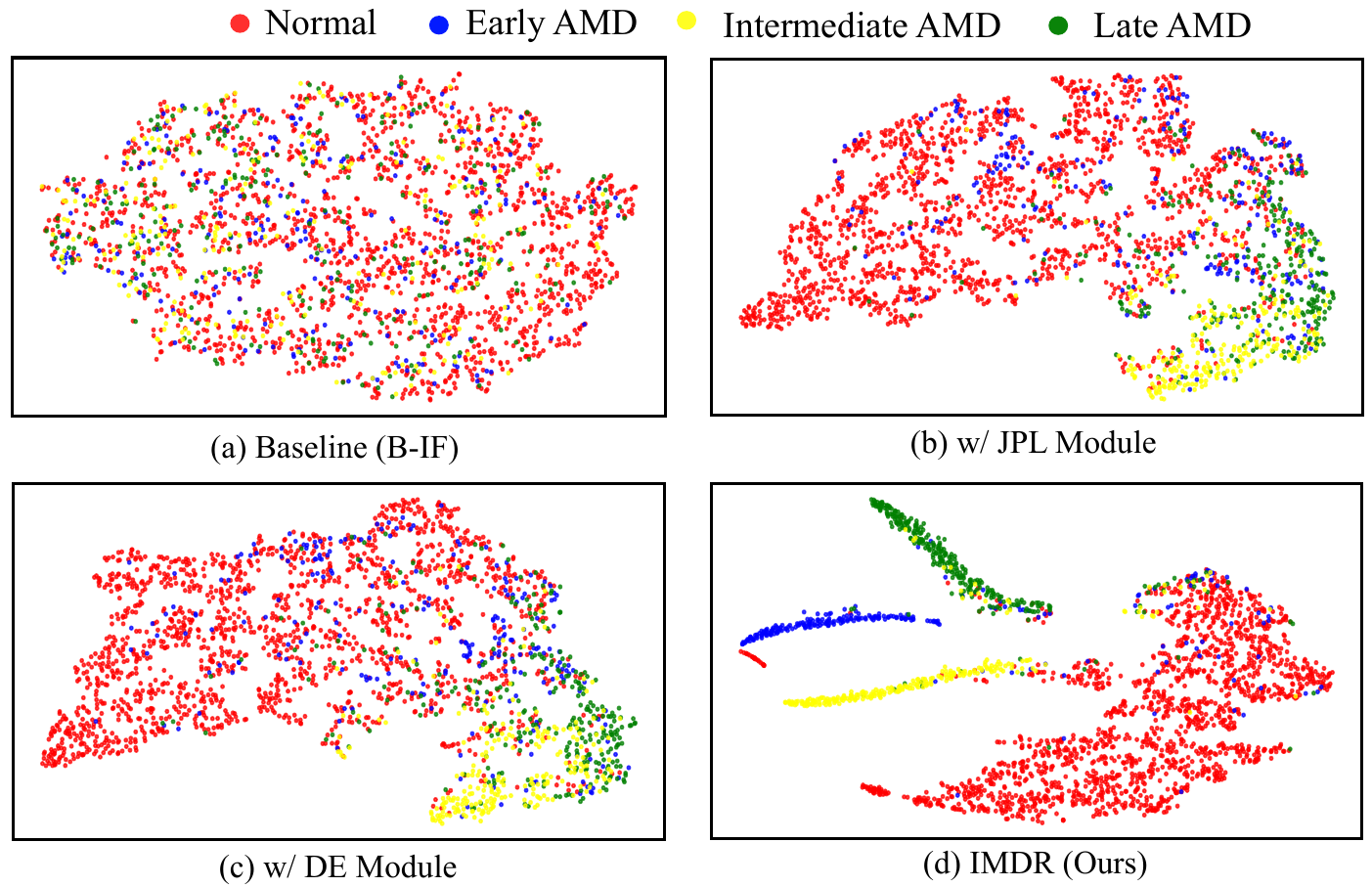}
\caption{Ablation Study of visualization under condition of missing OCT modality on Harvard-30k AMD test set.}
\label{fig: tsne}
\vspace{-0.5cm}
\end{figure}

To illustrate the effects of different modules, we perform a t-SNE visualization on 600 randomly selected samples from the AMD test set under missing OCT modality conditions. As shown in Fig.~\ref{fig: tsne} (a), the Baseline model struggles to extract distinct features from incomplete data. However, with the integration of JPL or DE modules, as shown in Fig.~\ref{fig: tsne} (b) and (c), features for Intermediate and Late AMD become more distinguishable, though some overlap remains. The model incorporating both JPL and DE modules achieves clear feature clustering within the same class and distinct separation between different classes, demonstrating the effectiveness of our disentangled representation strategy.

\subsubsection{Comparison with other distribution estimations.}
To validate the effectiveness of Joint Proxy Learning module in reducing intra-modality redundancy and enhancing the joint distribution, we assess the distribution estimation capabilities of various methods (Table~\ref{tb: dis}), as shown in Table~\ref{tb: dis}. Compared to PE~\cite{shi2019probabilistic} and PCME~\cite{chun2021probabilistic}, IMDR demonstrates a clear performance advantage, confirming the ability of the Joint Proxy Learning module to capture informative features, guide feature disentangling, and effectively handle severe modality incompleteness.

\begin{table}[t]
\centering
\normalsize 
\resizebox{0.3\textwidth}{!}{
\begin{tabular}{c|cccc}
\midrule
\textbf{Method} & \textbf{ACC} & \textbf{AUC} & \textbf{F1} \\
\midrule
C-Dist & 0.706  & 0.743 & 0.687 \\
PE & 0.717 & 0.770  & 0.706  \\
PCME & 0.735 & 0.789 & 0.713   \\
\rowcolor{gray!30} \textbf{IMDR (Ours)} & \textbf{0.752} & \textbf{0.804} & \textbf{0.766}   \\
\midrule
\end{tabular}}
\caption{Comparison with other distribution methods. C-Dist: directly estimates the distribution of features from each modality. PE: uses a fully connected layer to estimate the feature vector's distribution. PCME: incorporates attention modules to aggregate information from the feature map.}
\label{tb: dis}
\end{table}

\begin{figure}[t]
\centering
\includegraphics[width=\linewidth]{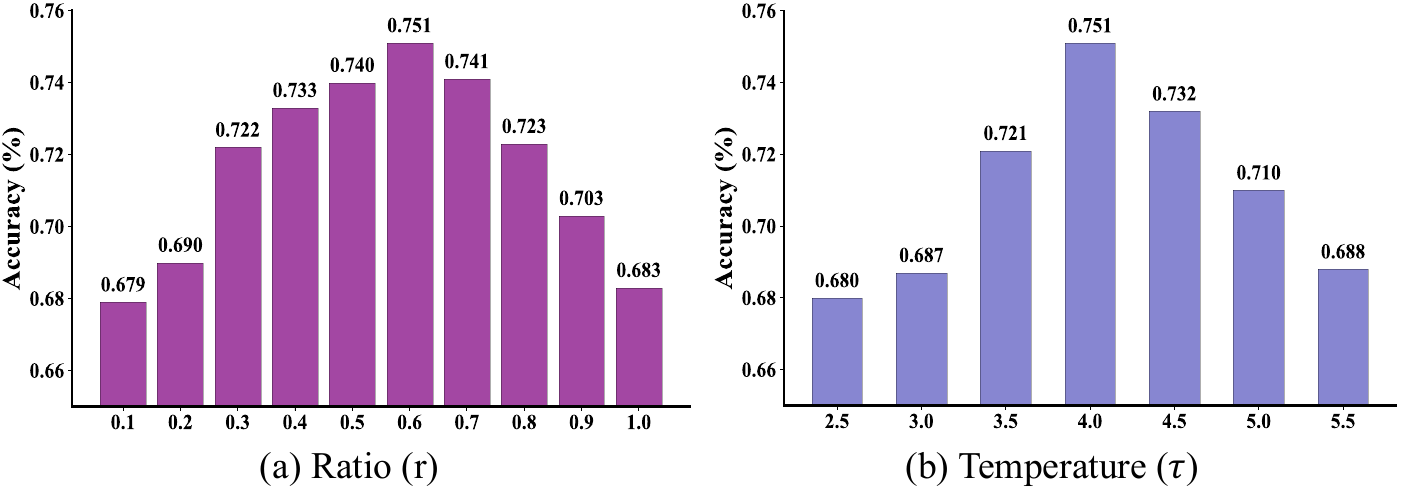}
    \caption{Ablation study of hyperparameters under the condition of missing OCT modality on Harvard-30k AMD test set. $r$: the ratio $r$ of number of proxy $N_p$ to the number of samples in the training set $N$. $\tau$: distillation temperature.}
\label{fig: sensitive}
\vspace{-0.5cm}
\end{figure}

\subsubsection{Sensitivity Analysis of Hyperparameters.}
We conduct a sensitivity analysis of key parameters within our IMDR model, as shown in Fig.~\ref{fig: sensitive}. The results indicate an initial increase of $r$ enhances model performance, then further increases lead to a decline due to noise from oversampling, which ultimately diminishes the overall effectiveness of our model.
Furthermore, as the $\tau$ coefficients increase, model performance initially improves but ultimately declines under higher parameter settings.

\section{Conclusions}
This paper identifies two critical limitations of existing methods: implicit representation constraints that limit capturing modality-specific information and modality heterogeneity, leading to feature distribution gaps and redundancy. To overcome it,  Incomplete Modality Disentangled Representation (IMDR) strategy disentangles features into distinct modal-shared and modal-specific components guided by mutual information. This enables student network to reconstruct missing semantics and produce robust multimodal representations. Additionally, we introduce a Joint Proxy Learning (JPL) module to eliminate intra-modality redundancy by leveraging class-specific proxies. Experiments on four ophthalmology multimodal datasets demonstrate that IMDR significantly outperforms state-of-the-art methods.

\bibliography{Main.bbl}

\end{document}